# The Optimality of Satisficing Solutions *

Othar Hansson and Andrew Mayer
Computer Science Department
University of California
Los Angeles, CA 90024

## Introduction

This paper addresses a prevailing assumption in single-agent heuristic search theory – that problem-solving algorithms should guarantee shortest-path solutions, which are typically called *optimal*. Optimality implies a metric for judging *solution quality*, where the *optimal* solution is the solution with the highest quality. When path-length is the metric, we will distinguish such solutions as *p-optimal*.

However, for most applications, finding p-optimal solutions requires exponential computation time. Shortest-path algorithms are then unacceptable to real-world users, who prefer *satisficing* solutions which can be achieved quickly.

Why are our problem-solving algorithms incompatible with the desires of real-world users? We contend that this is because the current approach to single-agent heuristic search has overlooked two important facts about real-world problem-solvers:

1. They use a variety of *attributes* to measure the quality of solutions, not simply path-length.

2. They are resource-bounded, and cannot perform the exponential computation required by shortest-path algoritms.

To address the first issue, we describe the application of multiattribute utility theory, a formal method for capturing the subjective preferences of a problem-solver, in the context of heuristic search. In doing so, we generalize the notion of solution quality, and therefore optimality, to incorporate user-defined attributes, both of solution-paths and the search algorithms which find them. We will refer to this as *s-optimality* (optimality over arbitrary, subjectively determined criteria). In such a framework, optimal solutions are those which are most preferred by, or most *satisfy* the user. P-optimality is then simply the special case in which the problem-solver's only concern is the length of the solution-path.

The second issue is the infeasibility of exponential computation. Such computation is only required to satisfy the *guarantee* of p-optimality. We offer theoretical and practical arguments against algorithms which make such guarantees, and in favor of probabilistic algorithms, which attempt to find solutions that maximize expected utility. We describe first results of research on this class of algorithms.

*Supported by a grant from Heuristicrats.

**Outline of the Paper**

The first section presents background information on current approaches to heuristic search in single-agent problems. Section 2 re-assesses existing algorithms' guarantee of p-optimality, and shows that guaranteeing either p-optimality or s-optimality is impossible for real-world problem-solvers. Because agents must therefore act without guarantees of optimality, Section 3 reviews decision-theoretic methods for rational decision-making under uncertainty. In Section 4, we discuss probabilistic algorithms which attempt to maximize solution quality while operating under uncertainty. Finally, Section 5 presents preliminary results and future directions of this research.

# 1 Background

## 1.1 Problem-Solving Search

The state-space approach to problem-solving considers a problem as a quadruple, $(S, O \subset S \times S, I \in S, G \subset S)$. $S$ is the set of possible *states* of the problem. $O$ is the set of *operators*, or transitions from state to state. $I$ is the one *initial state* of a problem instance, and $G$ is the set of *goal states*. Any problem can be represented as a state-space graph, where the states are nodes, and the operators are directed, weighted arcs between nodes (the weight associated with each operator, $O_i$, is its *cost* $C(O_i)$). Solving a single-agent problem consists of determining a sequence of operators, $O_1, O_2, \ldots O_n$ which, when applied to $I$, yields a state in $G$. Such a sequence is called a *solution-path*, with cost $\sum_{i=1}^{n} C(O_i)$. Without loss of generality, we equate cost with distance and will refer to a minimal cost solution-path as a *shortest-path*.

## 1.2 Traditional Algorithms

Many common algorithms (e.g. $A^*$) guarantee p-optimality. To do so, they must both find a solution-path and verify that no shorter one exists.

In theory, p-optimal solutions may be found by brute-force search over the state-space. However, as most problems' state-spaces are prohibitively large, current algorithms use heuristic methods to focus a search. To do so, such algorithms demand *admissible* heuristics, those which do not overestimate path-length. Despite improved performance over brute-force methods, these algorithms still require exponential computational time [6].

## 1.3 Non-Traditional Algorithms

Early researchers recognized that p-optimal solutions are seldom preferred to solutions which provide slightly longer paths but require less computational effort – these were known as satisficing solutions. This prosaic recognition, however, has led to few concrete results.

One research direction was to study the effects of using inadmissible heuristics within existing shortest-path algorithms. Though p-optimality could not be guaranteed, solution-paths of acceptable length could often be found quickly [6,7]. Unfortunately, there were few precise guarantees regarding the length of solution-paths which one could expect to find, or the computational effort required to do so.

Another approach considered real-world domains which require *satisficing search* [9]. Satisficing search is applied to problems where many solution-paths are deemed "good enough", and one seeks any such path while attempting to minimize computation time. In theorem-proving, for example, any proof is satisficing, and one wishes only to determine whether a proof exists.

Despite these efforts, most current research still centers on the development of techniques which guarantee p-optimal solutions – we will refer to this as the *guaranteed optimality* requirement on algorithms.

# 2 Generalized Optimality

Consider, however, a general evaluating a battle plan. While his overall objective is to win the battle, he will prefer some plans to others based on attributes such as danger to equipment, danger to civilians, proximity of journalists, etc. Another general might choose different attributes by which to measure the qual-ity of possible plans. Even if the two use the same attributes, they may weight their importance differently. Thus, an *optimal* solution to this problem may be different for each of these problem-solvers, as assessing the quality of solutions involves subjectivity.

Striving for p-optimality oversimplifies real-world problem-solving, because it optimizes only path-length. To determine the quality of a solution in the general case, we must consider the tradeoffs among all attributes the problem-solver deems relevant. We will refer to this extended view of optimality as *s-optimality* (optimality over arbitrary, subjectively determined criteria).

Continuing the example, assume that the general does not yet have a plan to evaluate. His problem is to choose, from among his staff of colonels, one who will construct a plan. One young colonel is extremely imaginative, and assures the general that he can produce a brilliant and safe strategy, but will require 3 days to prepare it. Another, a quick thinker, can formulate a plan in less than 3 hours, but admits that it will be somewhat risky. Had these plans been drawn up *a priori*, the general would almost certainly choose the former. Unfortunately, simply choosing between the plans is not the problem the general must solve. His task is to choose between the planners, the colonels. In situations where time is important enough to the general, he would be "forced" to opt for the second colonel and his quick and dirty plan. In this example, the complexity of developing a plan (*computation time*) is a crucial attribute to consider in assessing the plan's overall quality.

Thus solution quality is measured by user-defined attributes, which can describe both the ultimate solution-path and costs incurred by the algorithm which generated it. The algorithm (colonel), together with the solution-path (battle-plan) it generates, form a *solution* for the problem-solver (general). If in searching for a solution-path an algorithm drains resources which the problem-solver values, such as time or space, then the execution of the algorithm *degrades* the quality of the solutions it will provide. This will be shown to drastically alter our views about problem-solving algorithms.

## 2.1 Guarantees

Returning briefly to p-optimality, if an algorithm guarantees a p-optimal solution, it must both find it, and verify that no shorter solution exists.

Clearly, for brute-force algorithms this requires exponential time, to explore the tree of possible solution-paths. Heuristic methods are more efficient because they prune entire sub-trees (i.e. many paths) at once. Typically, however, pruning techniques are unable to reduce the exponential complexity of the verification process.

Having seen that guaranteeing p-optimality requires exponential time, we now consider the complexity of guaranteeing s-optimality.

## 2.2 An Impossibility and a Paradox

Consider three types of agents who may engage in problem-solving. The first is a typical real-world problem-solver who must operate with bounded resources. The second is an idealized problem-solver whose resources are unbounded. Constraints of the real-world do not restrict him, but, other things being equal, he prefers solutions which require fewer of his resources. The third agent is also idealized, and differs from the second only in that, other things being equal, he is indifferent to the quantity of resourses he must allocate to solve a problem.

The third agent is of little interest to us. Because he is indifferent to computation time, for example, we may conclude that he is indifferent to ever finding a solution to his problem - an algorithm which never terminates would satisfy him! We do not consider this behavior goal-oriented, nor this agent a problem-solver.

**Resource-Bounded Agents** We now show that for the resource-bounded agent, achieving s-optimality is impossible.

Consider the example of a robot mailman, who must deliver packages in its office building during an 8 hour workday. If the building and number of packages are sufficiently large, then any strategy which involves a precomputed plan (to visit each office, each day) will fail, because the plan will require more than 8 hours to execute. Similarly unacceptable is any algorithm which requires more than 8 hours to compute a delivery route. This problem-solver is constrained by the upper limit of 8 hours on computation and execution time.

Unfortunately, as problem size increases, the exponential complexity of verifying the optimality of a solution will exceed the robot's allotted time. Thus we conclude that for sufficiently large problems, a resource-bounded agent cannot guarantee s-optimality.

**Ideal Agents** For the ideal agent, who seeks to conserve his resources, achieving s-optimality yields a paradox.

Consider an algorithm $A$ which claims to guarantee s-optimality for any problem instance. Assume that the s-optimal solution to this problem is $X$, with solution-path $X_p$. One of the attributes of $X$ is the time $X_{tc}$ required to compute it. Computation time will be measured in terms of the critical operation of $A$ – node generation (generating a neighboring state, given a state and an operator). We assume that the algorithm can only know the effects of an operator by applying it – any other consideration of the effects of an operator constitutes node generation as well.

In order to find $X$, algorithm $A$ must generate *at least* the nodes in $X_p$. Otherwise, $A$ could not produce the complete path $X_p$, and would not be s-optimal.

On the other hand, $A$ may generate *at most* those nodes in $X_p$. If not, another solution, $X'$, which generates only those nodes in $X_p$, would be preferred, and thus $X$ could not be s-optimal.

Therefore, to achieve s-optimality, $A$ must expand *exactly* those nodes in $X_p$. However, if so, $A$ would have insufficient information to *verify* that $X$ was s-optimal: in particular, it could have missed a shorter path $X_{p'}$.

Consider that at any point in the search, $A$ is forced to choose a move based on the examination of only one operator – the correct one. If $A$ did not know the solution-path *a priori*, it could not guarantee that alternate moves might not have led to a shorter solution-path.

Paradoxically, if one attempts to verify that a solution is s-optimal, he degrades its quality, guaranteeing that it will not be s-optimal. Only by prior knowledge of the s-optimality of a solution could the agent have made such a guarantee. We conclude that even ideal agents cannot guarantee s-optimality.

An analogy can be made to Heisenberg's Uncertainty Principle, which states that an observer is not independent of the observations he makes, and that the act of observation disturbs the environment. Similarly, solution-paths do not exist independent of the algorithms which find them – the price of executing search algorithms is exacted from the solutions they yield. The paradox suggests that current approaches to heuristic search may be misguided.

### 2.3 Relaxing Guaranteed Optimality

In addition to theoretical paradoxes, there are many practical illustrations of the problems incurred when attempting to guarantee optimality:

**efficiency:** Algorithms which guarantee p-optimality are prohibitively expensive, and are therefore rarely used in the real-world. The best-known shortest-path algorithms require *hours* of processing to solve even a child's Fifteen Puzzle.

**applicability:** Many real-world problems impose constraints on the time an agent may hesitate between moves. Researchers who have studied two-player games have developed algorithms (e.g. Iterative Deepening) to cope with these constraints. In single-agent domains, researchers have failed, until recently [4], to do so.

**visibility:** Observing the consequences of intermediate actions is likely to increase an agent's understanding of its environment. This is the motivation apparent in many human activities, such as exploratory surgery.

**adaptability:** Unanticipated changes in an agent's environment may render pre-computed solutions useless. This could be

151

due to natural causes, or to adversaries, who deliberately attempt to thwart the agent's plans.

The guaranteed optimality requirement has caused problems in practice as well as in theory, limiting the applicability of single-agent search methods to an extremely small set of problems.

Thus, a problem-solver, theoretically incapable of guaranteeing s-optimality, cannot be certain, before he takes an action, that it will lead to the s-optimal solution. Operating under uncertainty, an agent can only *estimate* the eventual consequences, or *outcomes*, which will result from his immediate actions, and make decisions based on his subjective assignment of solution quality to these outcomes.

Researchers studying single-agent problems have traditionally avoided this issue of decision-making under uncertainty for two reasons: they fail to look beyond p-optimality in measuring solution quality, and their algorithms are unable to commit to actions without having verified p-optimality. However, a formal method for decision-making under uncertainty can be found in the field of Decision Theory, the elements of which we now summarize in the context of heuristic search.

## 3 Decisions under Uncertainty

### 3.1 Expected Value

Returning to p-optimality, for simplicity, the first approach in the context of search would likely be to relinquish the *guarantee* of p-optimality, but strive to minimize expected path-length whenever possible. This is equivalent to stating that the proper objective of a rational agent is to maximize the expected value of the outcome (his expected return).

The following problem is a simple example of the decisions which rational agents are likely to face. Consider a choice among possible outcomes, as illustrated in Figure 1.

Outcome (1) promises a 55-move solution to our problem – although that may be longer than we would like, we would receive it with absolute certainty. Outcome (2), however, is probabilistic. Depending on the outcome of a coin flip we will receive either a 10-move solution (heads) or a 90-move solution (tails). Using the expected value decision rule, we would analyze this situation as follows. Outcome (1) yields a 55-move solution all the time, so its expected value is simply 55-moves. Outcome (2) would yield a 10-move solution with probability $\frac{1}{2}$ and a 90-move solution with probability $\frac{1}{2}$, and thus would have expected value $(10 \text{ moves} \cdot \frac{1}{2}) + (90 \text{ moves} \cdot \frac{1}{2}) = 50$ moves. Assuming we wish to minimize solution-length, we should gamble and choose outcome (2). Similarly, if another outcome, (3), had offered a solution of 45-moves, we would have chosen it rather than the gamble.

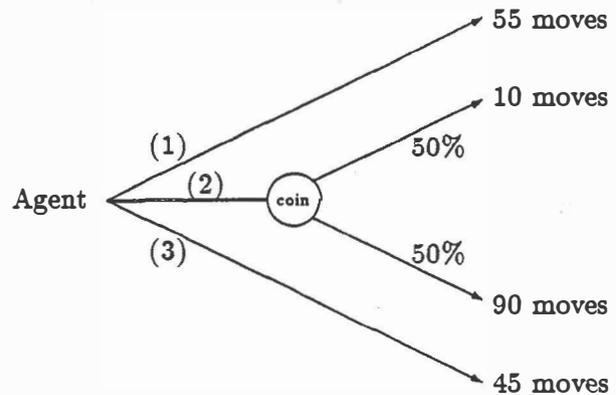

Figure 1: Agent to move

This decision rule, however, ignores the existence of overriding preferences which individual decision-makers may have. For example, faced with the choices outlined above, a *cautious* decision-maker might avoid the gamble in both situations.

### 3.2 Expected Utility

The theory of expected utility [10] was proposed as an improvement to the expected value rule, and has come to be the central tool in modern decision analysis. Utility is the subjective assignment of value to potential outcomes, when the exact outcome is uncertain. The theory claims that rational agents will attach utilities to all possible outcomes, and when faced with a decision under uncertainty, select that outcome with maximum expected utility. Maximizing expected utility thus subsumes the special case of maximizing expected value.

Note that the rational agent may be unable to produce such utility assignments on demand. But, if we subscribe to this theory, we can observe his decision-making, determine the utilities that he assigns to sample outcomes, and interpolate to produce his overall utility function [5].

It is important to note the inherent subjectivity of the utility assignments. For example, different decision-makers may have different attitudes towards the risk of uncertain outcomes. Recalling our earlier example (Figure 1), we are faced with a choice between two possible outcomes (1) & (2). Using the expected utility decision rule, our analysis changes. The expected utilities are as follows:

(1) $\qquad u(55 \text{ moves})$
(2) $\quad (u(10 \text{ moves}) \cdot \frac{1}{2}) + (u(90 \text{ moves}) \cdot \frac{1}{2})$

Ultimately, the choice will be based on the decision-maker's utility function $u()$.

For example, one might assign these utilities:

$$u(10 \text{ moves}) = 1.0$$
$$u(55 \text{ moves}) = 0.6$$
$$u(90 \text{ moves}) = 0.0$$

Here, the decision-maker desires the 10-move solution, but would prefer to avoid the possibility of receiving the 90-move solution. The 55-move solution is good enough. Based on his expected utility, he chooses outcome (1), avoiding disaster. His behavior is termed *risk averse*.

A *risk prone* decision-maker may be desperately trying to obtain a short solution. He agrees that a 10-move solution is superior, but sees little difference between the 55- and 90-move solutions, and assigns $u(55 \text{ moves}) = 0.1$. Based on his expected utility, he chooses the gamble, outcome (2).

### 3.3 Multiattribute Utility

Of course, real-world decision-makers are not faced with such simple choices, but typically consider utilities of many attributes, such as solution length, computation time, monetary cost, and memory usage. Required is a technique of assessing the tradeoffs among the many possible attributes of solutions.

The extension of utility theory which describes the behavior of a decison-maker faced with multiple, and possibly conflicting objectives is multiattribute utility theory. This theory allows one to combine utility functions of individual attributes into a joint utility function. A formal, rigorous presentation is offered in [5].

Oversimplifying, the techniques involve assessing the decision-maker's marginal utility of improving each of the attributes. In addition, the utility independence relations of the attributes must be determined by assessing whether the utility function for each attribute is independent of the values of the others.

Given this information, which can be elicited systematically from the decision-maker, one can determine what form the multiattribute function should take. If all the attributes are mutually utility independent, an additive function is used, as the decision-maker seeks simply to maximize the sum of the utilities. If not, a multiplicative or multilinear combination of the individual utility functions may be required. In the worst case of total interdependence one can only determine the decision-maker's utility function by observing his decisions and plotting points in $n$-dimensional attribute-space.

Once constructed, the multiattribute utility function can be evaluated for all potential outcomes, which are specified as an $n$-tuple of individual attribute values.

In summary, multiattribute utility theory offers a means for coping with uncertainty and subjectivity. Solution quality is determined by a user-defined utility function, which encapsulates the preferences of a problem-solver for different possible outcomes. Provided that we can determine the probability of each outcome, we can make decisions based on expected utility. In the next two sections, we discuss how these basic techniques can be realized, and applied in algorithms.

## 4 Algorithms

### 4.1 Off-Line and On-Line Algorithms

Current search algorithms for single-agent domains are unable to help resource-bounded problem-solvers solve real-world problems, because they fail to control the combinatorial explosion of state-spaces.

Because they must guarantee p-optimality, current algorithms are prevented from committing to even a single action before the ultimate goal is found. We term this static design process, in which computation wholly precedes execution, *off-line* problem-solving.

The uncertainty inherent in goal-oriented problem-solving demands that we develop algorithms which are capable of deciding on an action after only limited search. Such algorithms exemplify the characteristics of most intelligent real-world agents, who interact with their environment in the following manner:

**Sense:** Collect data about the surrounding environment using heuristic evaluation functions.

**Interpret:** Convert raw heuristic data into estimates of the probabilities of outcomes.

**Act:** Choose and apply the best operator.

This simple loop describes the behavior of resource-bounded agents who must operate in uncertain or complex environments.

We term this dynamic, interactive approach, which interleaves computation and execution, *on-line* problem-solving. An on-line algorithm commits to action without the luxury of certainty, potentially sacrificing optimality. It should be apparent that off-line algorithms, such as $A^*$, are merely that special case of on-line algorithms where the loop is executed only once. This may be appropriate if the problem solver has unbounded resources, and the length of the solution-path is the only attribute he seeks to optimize.

On-line algorithms are far more flexible than the off-line subclass. They can adapt to changing or hostile environments, incorporate new information after each move, make decisions under time and resource constraints, and produce satisficing solutions to problems whose p-optimal solutions are intractable.

A simple on-line algorithm is Minimin [4]. Minimin performs fixed-depth lookahead in a one-player game, and makes one move toward that leaf node with minimum heuristic value. Minimin follows the "face-value principle" of heuristic interpretation, ignoring the uncertainty of the information it receives from lookahead, and attempting no probabilistic assessment of solution quality. In decision-making under uncertainty, probabilistic assessment of outcomes is a prerequisite for the use of utility theory.

Sophisticated techniques for dealing with uncertainty are required. A system is under development which views heuristic evaluation functions as uncertain information regarding the outcome associated with each problem state [2], and exploits the constraints among these outcomes to better estimate the quality of the problem-solver's immediately available options [3].

### 4.2 One- and Two-Player Games

Research on single-agent problems and two-player games has bifurcated precisely on the central issues of this paper – computation time and optimality.

This can be understood by examining the origins of single-agent problem-solving research in the branch-and-bound techniques of combinatorial optimization, and the origins of two-player game research in Shannon's adaptation of game theory to chess [8]. Researchers in single-agent problem-solving designed algorithms which guaranteed p-optimal solutions because the problems they typically experimented with (such as the Eight Puzzle) were extremely simple.

In contrast, the obvious intractability of searching the complete chess tree led early researchers of two-player games to relinquish the guarantee of optimality afforded by the techniques they borrowed from game theory. Realizing the infeasability of off-line approachs, they employed on-line algorithms for two-player games. We suggest that single-agent problems require on-line algorithms as well.

Interestingly, the on-line approach discussed in [3] eliminates many of the traditional distinctions between single-agent and two-player domains: the differences are localized to the outcome constraints. In addition, the multiattribute utility formalism, and the utility-based decision-making mechanisms discussed here in the context of single-agent problems, easily extend to the two-player domain.

### 4.3 Utility of Algorithms

The generalized view of solution quality presented in Sections 2 and 3 suggests that algorithms should be judged based on the utility of the solutions they provide. The examples in Section 3 depicted a decision-maker faced with a choice of 2 possible outcomes. A problem-solver would find himself in an identical situation if he were forced to choose between the following two algorithms $X$ and $Y$. $X$ guarantees a solution which requires 35 minutes to compute but costs \$7500 to execute. $Y$ guarantees a solution which requires 2 weeks to compute and costs only \$1250 to execute. Problem-solvers should choose the better *algorithm* in precisely the same manner as decision-makers would choose the better *outcome*.

## 5 Optimizing Expected Utility: First Results

A problem-solver's *a priori* knowledge of the solutions provided by algorithms will rarely be so exact. Furthermore, current algorithms are not designed to predict the solutions that they will provide. However, a user can attempt to create performance models of algorithms. Having estimated the solutions that different algorithms are likely to provide, the user can determine the expected utility of each algorithm, and choose the best one for his purposes.

### 5.1 Performance Modeling

A practical technique for modeling the performance of on-line search algorithms is detailed in [1], which describes a system for automating the process. The system, with a problem instance and a user's multiattribute utility function as input, chooses, from among a given family of algorithms, the one which is expected to provide the user with the highest quality solutions. To estimate these outcomes, the system consults internal performance models which are based on a Markov model of the search process.

This system has been succesfully tested on the popular Eight Puzzle, where it optimized the performance of the Minimin algorithm for a user's utility function (with attributes of solution length, space and time). The level of lookahead is the only distinguishing parameter within the Minimin family of algorithms.

The multiplicative utility function used in the experiments was based on a hypothetical application, where elapsed time is limited to 10 minutes, memory to 10 megabytes, and solutions to 100 moves. The utility function is best summarized by three solutions that it judged to be relatively equivalent (note that while minimizing time and path-length are desirable, space is considered as a free, but bounded resource).

| PATH-LENGTH | SPACE | TIME |
| --- | --- | --- |
| 20 moves | $\leq$ 9 megabytes | 8 min. |
| 68 moves | $\leq$ 9 megabytes | 6 min. |
| 93 moves | $\leq$ 9 megabytes | 4 min. |

The system estimated, for a given problem instance, the performance of Minimin for different levels of lookahead, and output that lookahead level which maximized the user's expected utility. For example, for puzzles of depth $d = 19$, the system chose a lookahead level of 16 based on the performance model. Tested on 1000 puzzles of depth $d = 19$, that choice of lookahead provided the highest actual utility (compared with other lookahead levels).

Over all puzzle depths (1000 puzzle instances for each), the system's choices of lookahead level provided the highest actual utility 88.3% of the time, were within one level of the optimal 95.4% of the time, and never erred by more than three levels. Furthermore, in these worst cases, it provided solutions only slightly inferior to the best attainable – differing from the optimal choices by less than 0.1% in actual utility.

While the performance model and the Minimin algorithm are very simple, these successful results suggest that realistic performance estimates of search algorithms are attainable and powerful. Coupled with utility functions, they allow us to choose among algorithms, or choose parameters for algorithms, which maximize expected utility for the user.

### 5.2 Utility-Based Algorithms

Attempting to optimize expected utility by parameterizing existing algorithms is only an intermediate step. Real-world problem-solving requires algorithms which can more flexibly adapt to particular domains, and to particular users' utilities.

For example, a fixed lookahead level (as in Minimin) is a rigid, and unnecessary requirement for search algorithms. Determining the lookahead level dynamically (before each move) would be more flexible. The effort devoted to each decision would then be dependent on its estimated difficulty.

Algorithms could ultimately consider the utility of every action they take, including node generation, heuristic evaluation, etc., subject to the constraint that the time required to consider these factors is also a measure of efficiency. In such a setting, lookahead would examine some finite connected component of the state-space (rather than a fixed-depth full-width search tree), until the expected utility of searching further became less than the expected utility of committing to a move. An algorithm employing this approach is under development, as an extension of the system described in [3].

### Conclusion

This paper has addressed the guaranteed p-optimality requirement which pervades the study of single-agent problem-solving. We have found that p-optimality, given its computational cost, is usually an undesirable property, and that guaranteeing p-optimality is usually an impossible task, for resource-bounded agents.

Instead, such agents should be able to define their own measures of solution quality, and assign their own utilities to these solutions. Such subjective assessments of solution quality imply a subjective optimality, which can be formalized within multiattribute utility theory.

Toward this end, we have drawn a distiction between off-line and on-line approaches to problem-solving. On-line algorithms can operate under uncertainty, and within resource bounds, searching for solutions that maximize a problem-solver's expected utility.

We have demonstrated that considerable success can be achieved by parameterizing a simple existing algorithm based on the solution quality predictions of a performance model. We outline more sophisticated algorithms, equally applicable to both single-agent and two-player games, which are under development.

In summary, we depart from previous approaches, and propose that the important question for heuristic search theory to address is how to design real-world problem-solving systems which can allocate a user's bounded resources to best achieve his goals.

### Acknowledgements

We thank especially Moti Yung for his many insightful comments on this work. In addition, we would like to thank Rich Korf, Beth Mayer, and Risto Miikkulainen who commented on earlier drafts of this paper.